%% file: 0-main.tex
\documentclass{article}

\usepackage{graphicx} % more modern
\usepackage{subfigure} 

\usepackage{natbib}

\usepackage{hyperref}

\usepackage{booktabs} % For formal tables

\usepackage{enumitem}
\usepackage{float}
\usepackage{tabu}

\newtheorem{theorem}{Theorem}[section]

\newtheorem{definition}[theorem]{Definition}

\usepackage[accepted]{icml2013} 
\icmltitlerunning{Machine Teaching}

\begin{document} 

\twocolumn[
\icmltitle{Machine Teaching \\ 
           A New Paradigm for Building Machine Learning Systems}

% unless you've provided the [accepted] option to the icml2013
\icmlauthor{Patrice Y. Simard}{patrice@microsoft.com}
\icmlauthor{Saleema Amershi}{samershi@microsoft.com}
\icmlauthor{David M. Chickering}{dmax@microsoft.com}
\icmlauthor{Alicia Edelman Pelton}{aliciaep@microsoft.com}
\icmlauthor{Soroush Ghorashi}{sorgh@microsoft.com}
\icmlauthor{Christopher Meek}{meek@microsoft.com}
\icmlauthor{Gonzalo Ramos}{goramos@microsoft.com}
\icmlauthor{Jina Suh}{jinsuh@microsoft.com}
\icmlauthor{Johan Verwey}{joverwey@microsoft.com}
\icmlauthor{Mo Wang}{mowan@microsoft.com}
\icmlauthor{John Wernsing}{johnwer@microsoft.com}
\icmladdress{Microsoft Research, One Microsoft Way, Redmond, WA 98052, USA}

% the ``keywords" metadata in the PDF but will not be shown in the document
\icmlkeywords{machine teaching, machine learning,
programming, software engineering, concept decomposition, teaching principles, teaching process, democratization of machine learning, featuring}

\vskip 0.3in
]

\begin{abstract} 
The current processes for building machine learning systems require practitioners with deep knowledge of machine learning. This significantly limits the number of machine learning systems that can be created and has led to a mismatch between the demand for machine learning systems and the ability for organizations to build them. We believe that in order to meet this growing demand for machine learning systems we must significantly increase the number of individuals that can teach machines. We postulate that we can achieve this goal by making the process of teaching machines easy, fast and above all, universally accessible.

While machine learning focuses on creating new algorithms and improving the accuracy of ``learners", the machine teaching  discipline focuses on the efficacy of the ``teachers". Machine teaching as a discipline is a paradigm shift that follows and extends principles of software engineering and programming languages. We put a strong emphasis on the teacher and the teacher's interaction with data, as well as crucial components such as techniques and design principles of interaction and visualization. 

In this paper, we present our position regarding the discipline of machine teaching and articulate fundamental machine teaching principles. We also describe how, by decoupling knowledge about machine learning algorithms from the process of teaching, we can accelerate innovation and empower millions of new uses for machine learning models.
\end{abstract} 

\input{0-mainbody}

\end{document}

%% file: 0-mainbody.tex
\input{1-introduction}
\input{2-newdiscipline}

\input{3-analogy}
\input{4-teachers}

\input{5-teachingprocess}
\input{6-conclusion}
\input{7-acknowledgements}

% Bibliography
\bibliographystyle{ACM-Reference-Format}
\bibliography{8-references}

%% file: 1-introduction.tex
\section{Introduction}

The demand for machine learning (ML) models far exceeds the supply of ``machine teachers" that can build those models. Categories of common-sense understanding tasks that we would like to automate with computers include interpreting commands, customer support, or agents that perform tasks on our behalf. The combination of categories, domains, and tasks leads to millions of opportunities for building specialized, high-accuracy machine learning models. For example, we might be interested in building a model to understand voice commands for controlling a television or building an agent for making restaurant reservations. The key to opening up the large space of solutions to is to increase the number of machine teachers by making the process of teaching machines easy, fast and universally accessible.
	
A large fraction of the machine learning community is focused on creating new algorithms to improve the accuracy of the ``learners" (machine learning algorithms) on given labeled data sets. The machine teaching (MT) discipline is focused on the efficacy of the teachers \textit{given the learners}. The metrics of machine teaching measure performance relative to human costs, such as productivity, interpretability, robustness, and scaling with the complexity of the problem or the number of contributors. 

Many problems that affect model building productivity are not addressed by traditional machine learning. One such problem is concept evolution, a process in which the teacher's underlying notion of the target class is defined and refined over time \cite{kulesza2014structured}. Label noise or inconsistencies can be detrimental to traditional machine learning because it assumes that the target concept is fixed and is defined by the labels. In practice, concept definitions, schemas, and labels can change as new sets of rare positives are discovered or when teachers simply change their minds. 

Consider a binary classification task for \textit{gardening web pages} where the machine learner and feature set is fixed. The teacher may initially label botanical garden web pages as positive examples for the gardening concept, but then later decide that these are negative examples. Relabeling the examples when the target concept evolves is a huge burden on the teacher. From a teacher's perspective, concepts should be decomposable into sub-concepts and the manipulation of the relationship between sub-concepts should be easy, interpretable, and reversible. At the onset, the teacher could decompose gardening into sub-concepts (that include botanical gardens) and label the web page according to this concept schema. 

In this scenario, labeling for sub-concepts has no benefits to the machine learning algorithm, but it benefits the teacher by enabling concept manipulation. Manipulation of sub-concepts can be done in constant time (i.e., not dependent on the number of labels), and the teacher's semantic decisions can be documented for communication and collaboration. Addressing concept evolution is but one example of where the emphasis on the teacher's perspective can make a large difference in model building productivity. 

Machine teaching is a paradigm shift away from machine learning, akin to how other fields in programming language have shifted from optimizing performance to optimizing productivity with the notions of functional programming, programming interfaces, version control, etc. The discipline of machine teaching follows and extends principles of software engineering and languages that are fundamental to software productivity. Machine teaching places a strong emphasis on the teacher and the teacher's interaction with data, and techniques and design principles of interaction and visualization are crucial components. Machine teaching is also directly connected to machine learning fundamentals as it defines abstractions and interfaces between the underlying algorithm and the teaching language. Therefore, machine teaching lives at the interaction of the human-computer interaction, machine learning, visualization, systems and engineering fields. The goal of this paper is to explore machine learning model building from the teacher's perspective. 

%% file: 2-newdiscipline.tex
\section{The need for a new discipline}

% \textcolor{red}{[Goal of this section is to demonstrate the need for changing the status-quo and define machine teaching]}

In 2016, at one of Microsoft's internal conferences (TechFest) during a panel titled ``How Do We Build and Maintain Machine Learning Systems?", the host started the discussion by asking the audience ``What is your worst nightmare?" in the context of building machine learning models for production. A woman raised her hand and gave the first answer:
\begin{quote}
  ``[...] Manage versions. Manage data versions. Being able to reproduce the models. What if, you know, the data disappears, the person disappears, the model disappears... And we cannot reproduce this. I have seen this hundreds of times in Bing. I have seen it every day. Like... Oh yeah, we had a good model. Ok, I need to tweak it. I need to understand it. And then... Now we cannot reproduce it. That is my biggest nightmare!''
\end{quote}

To put context to this testimony, we review what building a machine learning model may look like in a product group:
\begin{enumerate}
\item A problem owner collects data, writes labeling guidelines, and optionally contributes some labels. 
\item The problem owner outsources the task of labeling a large portion of the data (e.g., 50,000 examples).
\item The problem owner examines the labels and may discover that the guidelines are incorrect or that the sampled examples are inappropriate or inadequate for the problem. When that happens, GOTO step 1.
\item An ML expert is consulted to select the algorithm (e.g., deep neural network), the architecture (e.g., number of layers, units per layer, etc.), the objective function, the regularizers, the cross-validation sets, etc.
\item Engineers adjust existing features or create new features to improve performance. Models are trained and deployed on a fraction of traffic for testing.
\item If the system does not perform well on test traffic, GOTO step 1
% * <pysimard@gmail.com> 2017-07-16T15:03:22.043Z:
% 
% > JINA: Don't we do feature engineering?
% Is this now addressed by step 5?
% 
% ^ <gonzo.ramos@gmail.com> 2017-07-16T19:42:48.585Z:
% 
% so we should say goto step 1 or 5?
%
% ^.
\item The model is deployed on full traffic. Performance of the model is monitored, and if that performance goes below a critical level, the model is modified by returning to step 1.
% * <pysimard@gmail.com> 2017-07-16T15:04:17.476Z:
% 
% > JINA: Feature engineering?
% Is this now addressed by step 5?
% 
% ^.
% * <gonzo.ramos@gmail.com> 2017-07-17T02:43:05.872Z:
% 
% I have edited step 8 to make it more compact. It reads better without the ``hanging" 1
% 
% ^.

\end{enumerate}

An iteration through steps 1 to 6 typically takes weeks. The system can be stable at step 7 for months. When it eventually breaks, it can be for a variety of reasons: the data distribution has changed, the competition has improved and the requirements have increased, new features are available and some old features are no longer available, the definition of the problem has changed, or a security update or other change has broken the code. At various steps, the problem owner, the machine learning expert, or the key engineer may have moved on to another group or another company. The features or the labels were not versioned or documented. No one understands how the data was collected because it was done in an ad hoc and organic fashion. Because multiple players with different expertise are involved, it takes a significant amount of effort and coordination to understand why the model does not perform as well as expected after being retrained. In the worst case, the model is operating but no one can tell if it is performing as expected, and no one wants the responsibility of turning it off. Machine learning ``litter" starts accumulating everywhere. These problems are not new to machine learning in practice \cite{technicaldebt}. 

The example above illustrates the fact that building a machine learning model involves more than just collecting data and applying learning algorithms, and that the management process of building machine learning solutions can be fraught with inefficiencies. There are other forms of inefficiencies that are deeply embedded in the current machine learning paradigm. For instance, machine learning projects typically consist of a single monolithic model trained on a large labeled data set. If the model's summary performance metrics (e.g., accuracy, F1 score) were the only requirements and the performance remained unchanged, adding examples would not be a problem even if the new model errs on the examples that were previously predicted correctly. However, for many problems for which predictability and quality control are important, any negative progress on the model quality leads to laborious testing of the entire model and incurs high maintenance cost. A single monolithic model lacks the modularity required for most people to isolate and address the root cause of a regression problem. 

\subsection{Definitions of machine learning and machine teaching}
It is difficult to argue that the challenges discussed above are given a high priority in the world's best machine learning conferences. These problems and inefficiencies do not stem from the machine learning algorithm, which is the central topic of the machine learning field; they come from the processes that use machine learning, from the interaction between people and machine learning algorithms, and from people's own limitations.

To give more weight to this assertion, we will define the machine learning research field narrowly as:

\begin{definition}[Machine learning research] 
Machine Learning research aims at making the \textit{learner} better by improving ML algorithms.
\end{definition}

This field covers, for instance, any new variations or breakthroughs in deep learning, unsupervised learning, recurrent networks, convex optimization, and so on. 

Conversely, we see version control, concept decomposition, semantic data exploration, expressiveness of the teaching language, interpretability of the model, and productivity as having more in common with programming and human-computer interaction than with machine learning. These ``machine teaching" concepts, however, are extraordinarily important to any practitioners of machine learning. Hence, we define a discipline aimed at improving these concept as:

\begin{definition}[Machine teaching research] 
Machine teaching research aims at making the \textit{teacher} more productive at building machine learning models.
\end{definition}

We have chosen these definitions to minimize the intersection between the two fields and thus provide clarity and scoping. The two disciplines are complementary and can evolve independently. Of course, like any generalization, there are limitations. Curriculum learning \cite{Bengio:2009}, for instance, could be seen as belonging squarely in the intersection because it involves both a learning algorithm and teacher behavior. Nevertheless, we have found these definitions useful to decide what to work on and what not to work on.

% In order for machine teaching discipline to evolve independently of machine learning, machine teaching solutions require well-defined interfaces describing the inputs (e.g. training data and schema constraints) and outputs (e.g. a classification function) of machine learning algorithms. Through these interfaces, the machine teaching solution can be decoupled from the implementation details of the underlying machine learning algorithms and can leverage any machine learning algorithms that support these interfaces. \textcolor{orange}{Furthermore, as long as the featuring language in the teaching solution is expressive enough, teachers can successfully use any reasonable\footnote{\textcolor{red}{By reasonable, we require that (1) the algorithm returns a function that matches the training data if it can (it is consistent) and (2) the complexity of the algorithm's function space increases with the dimension of the feature space.}} ML algorithm.}

\subsection{Decoupling machine teaching from machine learning}

Machine teaching solutions require one or more machine learning algorithms to produce models throughout the teaching process (and for the final output). This requirement can make things complex for teachers. Different deployment environments may support different runtime functions, depending on what resources are available (e.g., DSPs, GPUs, FPGAs, tight memory or CPU constraints) or what has been implemented and green-lighted for deployment. Machine learning algorithms can be understood as ``compilers" that convert the teaching information to an instance of the set of functions available at runtime. For example, each such instance might be characterized by the weights in a neural network, the ``means" in K-means, the support vectors in SVMs, or the decisions in decision trees. For each set of runtime functions, different machine learning compilers may be available (e.g., LBFGS, stochastic gradient descent), each with its own set of parameters (e.g., history size, regularizers, k-folds, learning rates schedule, batch size, etc.) 

Machine teaching aims at shielding the teacher from both the variability of the runtime and the complexity of the optimization. This has a performance cost: optimizing for a target runtime with expert control of the optimization parameters will always outperform generic parameter-less optimization. It is akin to in-lining assembly code. But like high-level programming languages, our goal with machine teaching is to reduce the human cost in terms of both maintenance time and required expertise. The teaching language should be ``write once, compile anywhere", following the ISO C++ philosophy.  

Using well-defined interfaces describing the inputs (feature values) and outputs (label value predictions) of machine learning algorithms, the teaching solution can leverage any machine learning algorithms that support these interfaces. We impose three additional system requirements: 

\begin{enumerate}
\item The featuring language available to the teacher should be expressive enough to enable examples to be distinguished in meaningful ways (a hash of a text document has distinguishing power, but it is not considered meaningful). This enables the teacher to remove feature blindness without necessarily increasing concept complexity. 
\item The complexity (VC dimension) of the set of functions that the system can return increases with the dimension of the feature space. This enables the teacher to decrease the approximation error by adding features. 
\item The available ML algorithms must satisfy the classical definition of learning consistency \cite{vapnik2013nature}. This enables the teacher to decrease the estimation error by adding labeled examples.  
\end{enumerate}

The aim of these requirements is to enable teachers to create and debug any concept function to an arbitrary level of accuracy without being required to understand the runtime function space, learning algorithms, or optimization.

%% file: 3-analogy.tex
\section{Analogy to programming}
% \textcolor{red}{[Goal of this section is to demonstrates that Machine Teaching as a field is closer to programming and HCI than to machine learning. We can leverage history. This helps set expectations. Probability of success is much higher.]}

In this section, we argue that teaching machines is a form of programming. We first describe what machine teaching and programming have in common. Next, we highlight several tools developed to support software development that we argue are likely to provide valuable guidance and inspiration to the machine teaching discipline. We conclude this section with a discussion of the history of the discipline of programming and how it might be predictive of the trajectory of the discipline of machine teaching.

\input{3.1-commonalities}
\input{3.2-pavingway}
\input{3.3-trajectory}

%% file: 3.1-commonalities.tex
\subsection{Commonalities and differences between programming and teaching}

Assume that a software engineer needs to create a stateless \textit{target function} (e.g., as in functional programming) that returns value $Y$ given input $X$. While not strictly sequential, we can describe the programming process as a set of steps as follows:

\begin{enumerate}
\item The target function needs to be specified
\item The target function can be decomposed into sub-functions
\item Functions (including sub-functions) need to be tested and debugged
\item Functions can be documented
\item Functions can be shared
\item Functions can be deployed 
\item Functions need to be maintained (scheduled and unscheduled debug cycles)
\end{enumerate}

Further assume that a teacher wants to build a target \textit{classification} function that returns class $Y$ given input $X$. The process for machine teaching presented in the previous section is similar to the set of programming steps above. While there are strong similarities, there are also significant differences, especially in the debugging step (Table~\ref{tab:debugging}). 

% Table
\begin{table*}[ht!]
\begin{center}
\caption{Comparison of debugging steps in programming and machine teaching}
\label{tab:debugging}
\begin{tabular}{p{6cm}|p{8.5cm}}
  \toprule
  Debugging in programming   & Debugging in machine teaching\\
  \hline
  	(3) Repeat:                  & (3) Repeat: \\
      \hspace{0.5 cm} (a) Inspect   & \hspace{0.5 cm} (a) Inspect \\
      \hspace{0.5 cm} (b) Edit code & \hspace{0.5 cm} (b) Edit/add knowledge (e.g., labels, features, ...)\\
      \hspace{0.5 cm} (c) Compile   & \hspace{0.5 cm} (c) Train \\
      \hspace{0.5 cm} (d) Test      & \hspace{0.5 cm} (d) Test \\ 
  	
%    \begin{enumerate}[label=(\alph*)]
%      \item Inspect
%      \item Edit code
%      \item Compile
%      \item Test
%    \end{enumerate}
%    & 
%    \begin{enumerate}[label=(\alph*)]
%      \item Inspect
%      \item Edit/add knowledge (e.g., labels, features, ...)
%      \item Train
%      \item Test
%    \end{enumerate}\\
  \bottomrule
\end{tabular}
\end{center}
\end{table*}

In order to strengthen the analogy between teaching and programming, we need a machine teaching \textit{language} that lets us express these steps in the context of a machine learning model building task. For programming, the examples of languages include C++, Python, JavaScript, etc. which can be compiled into machine language for execution. For teaching, the language is a means of expressing teacher knowledge into a form that a machine learning algorithm can leverage for training.  Teacher knowledge does not need to be limited to providing labels but can be a combination of schema constraints (e.g., mutually exclusive labels for classification, state transition constraints in entity extraction\footnote{In an address recognizer, we might want to require that the zip code appears after the state. }), labeled examples, and features. Just as new programing languages are being developed to address current limitations, we expect that new teaching languages will be developed that allow the teacher to communicate different types of knowledge and to communicate knowledge more effectively.
% * <gonzo.ramos@gmail.com> 2017-07-16T19:43:46.199Z:
% 
% How should we introduce Viterbi?
% 
% ^.

%% file: 3.2-pavingway.tex
\subsection{Programming paving the way forward}

As we have illustrated in the previous sections, current machine learning processes require multiple people of different expertise and strong knowledge dependency among them, there are no standards or tooling for versioning of data and models, and there is a strong co-dependency between problem formulation, training and the underlying machine learning algorithms. Fortunately, the emerging discipline of machine teaching can leverage lessons learned from the programming, software engineering and related disciplines. These disciplines have developed over the last half century and addressed many analogous problems that machine teaching aims to solve. This is not surprising given their strong commonalities. In this section, we highlight several lessons and relate them to machine teaching.

\subsubsection{Solving complex problems}

The programming discipline has developed and improved a set of tools, techniques and principles that allow software engineers to solve complex problems in ways that allow for efficient, maintainable and understandable solutions. These principles include problem decomposition, encapsulation, abstraction, and design patterns. Rather than discussing each of these, we contrast the differing expectations between software engineers solving a complex problem and machine teachers solving a complex problem. One of the most powerful concepts that allowed software engineers to write systems that solve complex problems is that of decomposition. The next anecdote illustrates its importance and power.

We asked dozens of software engineers the following: 
\begin{enumerate}
\item Can you write a program that correctly implements the game Tetris?
\item Can you do it in a month?
\end{enumerate}

The answer to the first question is universally ``yes". The answer to the second question varies from ``I think so" to ``why would it take more than 2 days?". The first question is arguably related to the Church-Turing thesis which states that all computable functions are computable by a Turing machine. If a human can compute the function, there exists a program that can perform the same computation on a Turing machine. In other words, given that there is an algorithm to implement the Tetris game, most respectable software engineers believe they can also implement the game on whatever machine they have access to and in whatever programming language they are familiar with. 
% * <gonzo.ramos@gmail.com> 2017-07-16T19:44:36.935Z:
% 
% Then let's remove the reference to wikipedia...
% 
% ^ <pysimard@gmail.com> 2017-07-16T21:17:04.853Z:
% 
% Done.
%
% ^.
The answer to the second question is more puzzling. The state space in a Tetris game (informally the number of configurations of the pieces on the screen) is very large, in fact, far larger than can be examined by the software engineer. Indeed, one might expect that the complexity of the program should grow exponentially with the size of the representation of a state in the state space. Yet, the software engineers seem confident that they can implement the game in under a month. The most likely explanation is that they consider the complexity of the implementation and the debugging to be polynomial in both the representation of the state space and the input. 

Let us examine how machine learning experts react to similar questions asked about teaching a complex problem:

\begin{enumerate}
\item Can you teach a machine to recognize kitchen utensils in an image as well as you do?
\item Can you do it in a month?
\end{enumerate}

When these questions were asked to another handful of machine learning experts, the answers were quite varied. While one person answered ``yes" to both questions without hesitation, most machine learning experts were less confident about both questions with answers including ``probably", ``I think so", ``I am not sure", and ``probably not". Implementing the Tetris game and recognizing simple non-deformable objects seem like fairly basic functions in either fields, thus it is surprising that the answers to both sets of questions are so different. 

The goal of both programming and teaching is to create a function. In that respect, the two activities have far more in common than they have differences. In both cases we are writing functions, so there is no reason to think that the Church-Turing thesis is not true for teaching. Despite the similarities, the expectations of success for creating, debugging, and maintaining such function differ widely between software engineers and teachers.  While the programming languages and teaching languages are different, the answers to the questions were the same for all software engineers regardless of the programming languages. Yet, most machine learning experts did not give upper bounds on how long it would take to solve a teaching problem, even when they thought the problem was solvable.

Software engineers have the confidence of being able to complete the task in a reasonable time because they have learned to decompose problems into smaller problems. Each smaller problem can be further decomposed until coding and debugging can be done in constant or polynomial time. For instance, to code Tetris, one can create a state module, a state transformation module, an input module, a scoring module, a shape display module, an animation module, and so on. Each of these modules can be further decomposed into smaller modules. The smaller modules can then be composed and debugged in polynomial time. Given that each module can be built efficiently, software engineers have confidence that they can code Tetris in less than a month's time. 

It is interesting to observe that that the ability to decompose a problem is a learned skill and is not easy to learn. A smart student could understand and learn all the functions of a programming language (variables, arrays, conditional statements, for loops, etc.) in a week or two. If the same student was asked to code Tetris after two weeks, they would not know where to start. After 6 to 12 months of learning how to program, most software engineers would be able to accommodate the task of programming the Tetris game in under a month.

Akin to how decomposition brings confidence to software engineers\footnote{For similar reasons, the ability to decompose also bring confidence to professional instructors and animal trainers.} and an upper bound to solving complex problems, machine teachers can learn to decompose complex machine learning problems with the right tools and experiences, and the machine teaching discipline can bring the expectations of success for teaching a machine to a level comparable to that of programming.
% * <pysimard@gmail.com> 2017-07-16T21:24:41.737Z:
% 
% > For similar reasons, the ability to decompose also bring confidence to professional instructors and animal trainers
% I don't know if this is useful (?)
% 
% ^.

\subsubsection{Scaling to multiple contributors}

The complexity of the problems that software engineers can solve has increased significantly over the past half century, but there are limits to the scale of problems that one software engineer can solve. To address this, many tools and techniques have been developed to enable multiple engineers to contribute to the solution of a problem. In this section, we focus on three concepts - programming languages, interfaces (APIs), and version control.

One of the key developments that enables scaling with the number of contributors is the creation of standardized programming languages. The use of a standardized programming language along with design patterns and documentation enables other collaborators to read, understand and maintain the software. The analog to programming languages for machine teaching is the expressions of a teacher's domain knowledge which include labels, features and schemas. Currently, there is no standardization of the programming languages for machine teaching.

Another key development that enables scaling with the number of contributors is the use of componentization and interfaces, which are closely related to the idea of problem decomposition discussed above. Componentization allows for a separation of concerns that reduces development complexity, and clear interfaces allow for independent development and innovation. For instance, a software engineer does not need to consider the details of the hardware upon which the solution will run. For machine teaching, the development of clear interfaces for services required for teaching, such as training, sampling and featuring, would enable independent teaching. In addition, having clear interfaces for models, features, labels, and schemas enables composing these constituent parts to solve more complex problems, and thus, allowing for their use in problem decomposition.

The final development that enables scaling with the number of contributors is the development of version control systems. Modern version control systems support merging contributions by multiple software engineers, speculative development, isolation of bug fixes and independent feature development, and rolling back to previous versions among many other benefits. The primary role of a version control system is to track and manage changes to the source code rather than keeping track of the compiled binaries. Similarly, in machine teaching, a version control system could support managing the changes of the labels, features, schemas, and learners used for building the model and enable reproducibility and branching for experimentation while providing documentation and transparency necessary for collaboration.

\subsubsection{Supporting the development of problem solutions}

In the past few decades, there has been an explosion of tools and processes aimed at increasing programming productivity. These include the development of high-level programming languages, innovations in integrated development environments, and the creation of development processes. Some of these tools and processes have a direct analog in machine teaching, and some are yet to be developed and adapted. Table~\ref{tab:mapping} presents a mapping of many of these tools and concepts to machine teaching.

% Table
\begin{table*}[th!]
\begin{center}
\caption{Mapping between programming and machine teaching}
\label{tab:mapping}
\begin{tabu} to 0.8\textwidth { X[c] | X[c] }
  \toprule
  Programming   & Machine teaching\\
  \hline
Compiler & ML Algorithms (Neural Networks, SVMs)\\ \\
Operating System/Services/IDEs & Training, Sampling, Featuring Services, etc. \\ \\
Frameworks & ImageNet, word2vec, etc. \\ \\
Programming Languages (Fortran, Python, C\#) & Labels, Features, Schemas, etc. \\ \\
Programming Expertise & Teaching Expertise \\ \\
Version Control & Version Control \\ \\
Development Processes (specifications, unit testing, deployment, monitoring, etc.) & Teaching Processes (data collection, testing, publishing, etc.)\\
  \bottomrule
\end{tabu}
\end{center}
\end{table*}

%% file: 3.3-trajectory.tex
\subsection{The trajectory of the machine teaching discipline}

We conclude this section with a brief review of the history of programming and how that might inform the trajectory of the machine teaching discipline. The history of programming is inexorably linked to the development of computers. Programming started with scientific and engineering tasks (1950s) with few programs and programming languages like FORTRAN that focused on compute performance. In the 1960s, the range of problems expanded to include management information systems and the range of programming languages expanded to target specific application domains (e.g., COBOL). The explosion of the number of software engineers led to the realization that scaling with contributors was difficult \cite{brooks1995mythical}. In the 1980s, the scope of problems to which programming was applied exploded with the advent of the personal computer as did the number of software engineers solving the problems (e.g., with Basic). Finally, in the 1990s, another explosive round of growth began with the advent of web programming and programming languages like JavaScript and Java. As of writing this paper, the number of software engineers in the world is approaching 20 million!
% * <pysimard@gmail.com> 2017-07-16T22:22:42.905Z:
% 
% > In the 1960s, the range of problems expanded to include management information systems and, the range of programming languages expanded to target specific application domains (e.g., COBOL).
% I would like to add the reference to the Mythical Man Month, which was the painful realization that something needed to be done to scale with the number of programmers.
% 
% ^ <pysimard@gmail.com> 2017-07-17T01:22:00.462Z <jasmart@gmail.com> 2017-07-17T01:22:01.691Z.
% ^ <jasmart@gmail.com> 2017-07-17T01:22:17.771Z.

Machine teaching is undergoing a similar explosion. Currently, much of the machine teaching effort is undertaken by experts in machine learning and statistics. Like the story of programming, the range of problems to which machine learning has been applied has been expanding. With the deep-learning breakthroughs in perceptual tasks in the 2010s (e.g., speech, vision, self-driving cars), there has been an incredible effort to broaden the range of problems addressed by teaching machines to solve the problems. Similar to the expanding population of software engineers, the advent of services like LUIS.ai\footnote{\url{https://www.luis.ai/}} and Wit.ai\footnote{\url{https://wit.ai/}} have enabled domain experts to build their own machine learning models with no machine learning knowledge. The discipline of machine teaching is young and in its formative stages. One can only expect that this growth will continue at an even quicker pace. In fact, machine teaching might be the path to bringing machine learning to the masses.

%% file: 4-teachers.tex
\section{The role of teachers}

The role of the teacher is to transfer knowledge to the learning machine so that it can generate a useful model that can approximate a concept. Let's define what we mean by this. 

\begin{definition}[Concept] 
A concept is a mapping from any example to a label value.
\end{definition}

For example, the concept of a recipe web page can be represented by a function that returns zero or one, based on whether a web page contains a cooking recipe. In another example, an address concept can be represented by a function that, given a document, returns a list of token ranges, each labeled ``address", ``street", ``zip", ``state", etc.
% * <pysimard@gmail.com> 2017-07-16T15:00:25.492Z:
% 
% > cloud
% Remove? This seems like a spurious word
% 
% ^ <gonzo.ramos@gmail.com> 2017-07-17T01:46:15.190Z:
% 
% is can better?
%
% ^ <pysimard@gmail.com> 2017-07-17T03:02:31.628Z.
Label values for a binary concept could be ``Is" and ``Is Not". We may also allow a ``Undecided" label which allows a teacher to postpone labeling decisions or ignore ambiguous examples. Postponing a decision is important because the concept may be evolving in the teacher's head. An example of this is in \cite{kulesza2014structured}.

\begin{definition}[Feature]
A feature is a concept that assigns each example a scalar value.
\end{definition}

We usually use feature to denote a concept when emphasizing its use in a machine learning model. For example, the concept corresponding to the presence or absence of the word ``recipe" in text examples might be a useful feature when teaching the recipe concept.

\begin{definition}[Teacher]
A teacher is the person who transfers concept knowledge to a learning machine.
\end{definition}

To clarify this definition of a teacher, the methods of knowledge transfer need to be defined. At this point, they include a) example selection (biased), b) labeling, c) schema definition (relationship between labels), d) featuring, and e) concept decomposition (where features are recursively defined as sub-models). The teachers are expected to make mistakes in all the forms of knowledge transfer. These teaching ``bugs" are common occurrences.

\begin{figure*}
  \begin{center}
  \includegraphics[width=0.8\textwidth]{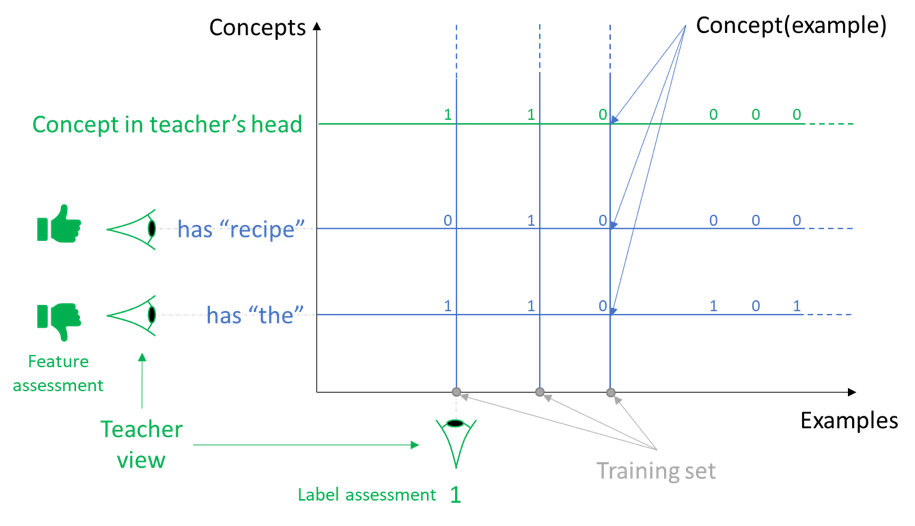}
  \end{center}
  \caption{Representation of examples and concepts. Each column represents an example and contains all concept values for that example. A teacher looks in that direction to ``divine" a label. The teacher has access to feature concepts not available to the training set (it is part of the teaching power). However, the teacher does not know his/her own program. Each row represents a concept and contains the value of that concept for all examples. A teacher looks in that direction to ``divine" the usefulness of a feature concept. A teacher can guess the values over the space of examples (it is part of the teaching power). Features selected by the teacher looking horizontally are immune to over-training. }
  \label{fig:concepts}
\end{figure*}

Figure~\ref{fig:concepts} illustrates how concepts, labels, features, and teachers are related. We assume that every concept is a computable function of a representation of examples. The representation is assumed to include all available information about each example. The horizontal axis represents the (infinite) space of examples. The vertical axis represents the (infinite) space of programs or concepts. In computer science theory, programs and examples can be represented as (long) integers. Using that convention, each integer number on the vertical axis could be interpreted as a program, and each integer number on the horizontal axis could be interpreted as an example. We ignore the programs that do not compile and the examples that are nonsensical. We now use Figure~\ref{fig:concepts} to refer to the different ways a teacher can pass information to a learning system.

\begin{definition}[Selection]
Selection is the process by which teachers gain access to an example that exemplifies useful aspects of a concept. 
\end{definition}

Teachers can select specific examples by filtering the set of unlabeled examples. By choosing these filters deliberately, they can systematically explore the space and discover information relevant to concepts. For example, a teacher may discover insect recipes while building a recipe classifier by issuing a query on ``source of proteins". We note that uniform sampling and uncertainty sampling, which have no explicit input from a teacher, are likely of little use for discovering rare clusters of positive examples. Combinations of semantic filters involving trained models are even more powerful (e.g., ``nutrition proteins" and low score with current classifier). This ability to find examples containing useful aspects of a concept enables the teacher to find useful features and provide the labels to train them. Furthermore, the selection choices themselves can be valuable documentation of the teaching process.

\begin{definition}[Label]
A label is a (example, concept value) pair created by a teacher in relation to a concept.
% * <pysimard@gmail.com> 2017-07-16T16:11:27.140Z:
% 
% > Labels
% Following Chris's suggestion, I am changing the definition of labels to be a pair rather than a concept value given a document. It is a bit risky because most people think of labels as concept values (e.g. 1 or 0). The concept values have no information on their own. The information is in the pairs. For the teacher to communicate knowledge through labels, the teacher need to provide pairs. Labels, for the purpose of this document, are always created by teachers.
% 
% ^ <pysimard@gmail.com> 2017-07-17T03:02:49.755Z.
\end{definition}

Teachers can provide labels by ``looking at a column" in Figure~\ref{fig:concepts}. It is important to realize that the teachers do not know which programs are running in their heads when they evaluate the target concept values. If they knew the programs, they would transfer their knowledge in programmatic form to the machine and would not need machine learning. Teachers instead look at the available data of an example and ``divine" its label. They do this by unconsciously evaluating sub-features and combining them to make labeling decisions. The feature spaces and the combination functions available to the teachers are beyond what is available through the training sets. This power is what makes the teachers valuable for the purpose of creating labels. 
%When teachers are asked about their label decision, they typically refer to sub-concepts that they can identify, e.g. ``this is a recipe because it contains the word ``recipe". In other words, they identify the recipe concept (concept = document contains the word ``recipe") and its value as part of an explanation. Some teachers are better than others at explaining their decisions. 
% * <pysimard@gmail.com> 2017-07-16T16:33:01.063Z:
% 
% > When teachers are asked about their label decision
% That actually is not related to labeling.
% 
% ^.

\begin{definition}[Schema]
A schema is a relationship graph between concepts.
\end{definition}
When multiple concepts are involved, a teacher can express relationship between them. For instance, the teacher could express that the concepts ``Tennis" and ``Soccer" are mutually exclusive, or that concept ``Tennis" implies the concept ``Sport". These concept constraints are relationships between lines on the diagram (true across all examples). Separating knowledge captured by the schema from the knowledge captured by the labels allows information to be conveyed and edited at a high level. The implied labels can be changed simply by changing the concept relationship. For instance, ``Golf" could be moved from being a sub-concept of ``Sport" to being mutually exclusive or vice versa. Teachers can understand and change the semantics of a concept by reviewing its schema. Semantic decisions can be reversed without editing individual labels.

\begin{definition}[Generic feature]
A generic feature is a set of related feature functions.
% * <pysimard@gmail.com> 2017-07-16T17:11:18.606Z:
% 
% > scalar value
% Addressing Jina's concern. I also introduce the implication that every feature is a concept. The converse is not true when the feature space is restricted (e.g. to dictionaries).
% 
% ^ <gonzo.ramos@gmail.com> 2017-07-16T18:51:17.899Z:
% 
% I used R here, I did not introduce it. I presumed that it is familiar enough to a math friendly audience... I like the use os scalar instead.
%
% ^.
\end{definition}

Generic features are created by engineers in parametrizable form, and teachers instantiate individual features by providing useful and semantic parameters. For instance, a generic feature could be: ``Log(1 + number of instances of words in list $X$ in a document)" and an instantiation would be setting $X$ to a list of car brands (useful for an automotive classifier). 

Given a set of generic features, teachers have the ability to evaluate different (instantiated) features by looking along the corresponding horizontal lines in Figure~\ref{fig:concepts}. Given two features, the teachers can ``divine" that one is better than the other on a large unlabeled set. For instance, a teacher may choose a feature that measures the presence of the word ``recipe" over a feature that measures the presence of the word ``the", even though the latter feature might yield better results on the training set. This ability to estimate the value of a feature over estimated distributions of the test set is essential to feature engineering, and is probably the most useful capability of a teacher. Features selected by the teacher in this manner are immune to over-training because they are created independently of the training set. Note the contrast to ``automatic feature selection", which only looks at the training set and concept-independent statistics and is susceptible to over-training.

\begin{definition}[Decomposition]
Decomposition is the act of using simpler concepts to express more complex ones. 
\end{definition}

Whereas teachers do not have direct access to the program implementing their concept, they sometimes can infer how these programs work. Socrates used to teach by asking the right questions. The ``right question" is akin to providing a useful sub-concept, whose value makes evaluating the overall concept easier. In other words, Socrates was teaching by decomposition rather than by examples. This ability is not equally available to teachers. It is learned. It is essential to scaling with complexity and with the number of teachers. It is the same ability that helps software engineers decompose functions into sub-functions. Software engineers also acquire this ability with experience.  As in programming, teaching decompositions are not unique (in software engineering, switching from one decomposition to another is called refactoring). 

The knowledge provided by the teacher through concept decomposition is high level and modular. Each concept implementation might provide its own example selection, labels, schema, and features. These can be viewed as documentation of interfaces and contracts. Each concept implementation may be a black box, but the concept hierarchy is transparent and interpretable. Concept decomposition is the highest form of knowledge provided by the teacher.

Now that we have defined some of the key roles of the (machine) teacher, we turn to the question of how do we meet the demand for them.

\input{4.1-who}

%% file: 4.1-who.tex
\subsection*{Meeting the demand for teachers}
We postulate that the right solution to satisfy the increasing demand for machine learning models is to increase the number of people that can teach machines these models. But how do we do that and who are they?

The current ML-focused work flows put the machine learning or data scientist on the driver's seat.  While training more scientists is a way to increase the number of teachers, we believe that that is not the right path to follow. For starters, machine learning and data scientists are a scarce and expensive resource. Secondly, machine learning scientists can serve a better purpose inventing and optimizing learning algorithms. In the same way, data scientists are indispensable applying their expertise to make sense of data and transform it into a usable form.

The machine teaching process that we envision does not require the skills of a ML expert or data scientist. Machine teachers use their domain knowledge to pick the right examples and counterexamples for a concept and explain why they differ. They do this through an interactive information exchange with a learning system. It is within the ranks of the domain experts where we will find the large population of machine teachers that will increase, by orders of magnitude, the number of ML models used to solve problems. We can transform domain experts by making a machine teaching language universally accessible. 

A key characteristic of domain experts is that they understand the semantics of a problem. To this point, we argue that if a problem's data does not need to be interpreted by a person to be useful, machine teaching is not needed. For example, problems for which the labeled data is abundant or practically limitless; e.g. Computer Vision, Speech Understanding, Genomics Analysis, Click-Prediction, Financial Forecasting. For these, powerful learning algorithms or hardware may be the better strategy to arrive at an effective solution. In other problems like the above, feature selection using cross validation can be used to arrive at a good solution without the need of a machine teacher.

There is nonetheless, an ever-growing set of problems for which machine teaching is the right approach; problems where unlabeled data is plentiful and domain knowledge to articulate a concept is essential. Examples of these include controlling Internet-of-Things appliances through spoken dialogs and the environment's context, or routing customer feedback for a brand new product of a start-up to the right department, building a one-time assistant to help a paralegal sift through hundreds of thousands of briefs, etc.

We aim at reaching the same number of machine teachers as there are software engineers, a set counted in the tens of millions. Table~\ref{tab:teachers} illustrates the differences in numbers between machine learning scientists, data scientists, and domain experts. By enabling domain experts to teach, we will enable them to apply their knowledge to solve directly millions of meaningful, personal, shared, ``one-off" and recurrent problems at a scale that we have never seen.
% * <gonzo.ramos@gmail.com> 2017-07-16T19:33:14.487Z:
% 
% I need help formatting this table. LATEX experts unite!
% 
% ^ <pysimard@gmail.com> 2017-07-17T01:06:05.225Z:
% 
% I think I fixed the random white lines. Is this what you wanted?
%
% ^ <gonzo.ramos@gmail.com> 2017-07-17T02:31:57.368Z:
% 
% Yes. Thank you!
%
% ^ <pysimard@gmail.com> 2017-07-17T02:59:44.510Z.
% Table
\begin{table*}[th!]
\begin{center}
\caption{Where to find machine teachers}
\label{tab:teachers}
\begin{tabu} to 0.9\textwidth {X[c]|X[c]|X[c]}
  \toprule
  Potential teacher & Quantities & Characteristics\\
  \hline
    Machine learning experts & Tens of thousands & Has profound understanding of machine learning. Can modify a machine learning algorithm or architecture to improve performance.\\
    \hline
    Data Scientist / Analyst & Hundreds of thousands & Can analyze big data, detect trend and correlations using machine learning. Can train machine learning models on existing values to extract value for a business.\\
    \hline
    Domain expert & Tens of millions & Understands the semantics of a problem. Can provide examples and counter examples, and explain the difference between them. \\
    \bottomrule
 \end{tabu}
\end{center}
\end{table*}

%% file: 5-teachingprocess.tex
\section{Teaching process}

% \textcolor{red}{[The purpose of this section is to 1) summarize the current machine learning processes, 2) propose new processes inspired by programming, 3) demonstrate that this is a better approach to machine learning than current approach.]}

A teaching or programming language can be applied in many different ways, some more effective than others. We propose the following principles for the language and process of machine teaching:

% \begin{principle}
\paragraph{Universal teaching language}
We do not rely on the power of specific machine learning algorithms. The teaching interface is the same for all algorithms. If a machine learning algorithm is swapped for another one, more teaching may be necessary, but the teaching language and the model building experience is not changed. Machine learning algorithms should be interchangeable. Conversely, the teaching language should be simple and easy to learn given the domain (e.g., text, signal, images). Ideally, we aim at designing an ANSI or ISO standard per domain. Teachers that speak the same language should be interchangeable.

\paragraph{Feature completeness (or realizability)}
We assume that all the target concepts that a teacher may want to implement are ``realizable" through a recursive composition of models and existing features. This implies a property on the feature set, which we call ``feature completeness". Feature completeness is the responsibility of the teaching tool, not the teachers. Teachers achieve realizability through the following actions: 

\begin{enumerate}
\item {\bf Add missing features:} If a teacher can distinguish two documents belonging to two different classes in a meaningful way, there must be a (corresponding) feature expressible in the system that can make an equivalent semantically meaningful distinction. By adding such a feature, the teacher can correct feature blindness errors. If no such feature exists, the language is not feature complete for distinguishing the desired classes.

\item {\bf Create features through decomposition:} If the concept function cannot be learned from the existing set of features due to limitations of the model class, the teacher can circumvent this problem by creating features that are themselves models; we call this process ``model decomposition". To illustrate the point, suppose there are two binary features $A$ and $B$, and the teacher would like to produce a model for $A \oplus B$ (where $\oplus$ stands for XOR) using logistic regression. Because of the capacity limitations of logistic regression, it is impossible to represent $A \oplus B$ without additional features. If the teacher adds a third AND feature $A \wedge B$, however, logistic regression can work. 
% ($A \oplus B = A + B - \left( A \wedge B\right) $). 
Note that $A \wedge B$ is itself learnable via logistic regression in the $A$ and $B$ feature space.  
%Note that a hash of the document is a feature than can distinguish any two documents, but it is not semantically meaningful.
%\item {\bf Insufficient capacity remedy:} If the concept function does not belong to the class of functions computable by a model given a current feature set, the teacher can add features that reduce the complexity of the problem in a new feature space. For example, consider two binary features $A$ and $B$. The function $A \oplus B$ (where $\oplus$ stands for XOR) is not computable using a logistic regression model. If a teacher adds a third AND feature $A \wedge B$, it becomes possible to train a logistic regression model to compute a function equivalent to $A \oplus B$ as a linear combination of 3 features ($A + B - \left( A \wedge B\right) $). Note that $A \wedge B$ is itself learnable via logistic regression in the $A$ and $B$ feature space. 
%In other words, the model capacity limitation in a given feature space can be  circumvented by composing models of low capacity in different augmented feature spaces. This ability is critical to learn complex concepts.

\item {\bf Explicitly ignore ambiguous patterns:} Ambiguous patterns can be marked as ``don't care" to avoid wasting features, labels, and the teacher's time on difficult examples. Areas of ``don't care" are used as a coping mechanism to keep the realizability assumption despite the Bayes error rate. This action does not constrain the feature set.  

\end{enumerate}

Feature completeness of a teaching language does not imply that the language can be used to efficiently teach concepts. If a feature complete language is not very expressive, realizability can require a large number of model compositions. If a feature complete language is too expressive (e.g. a features can be specified as programs), the teachers have to become engineers.

\paragraph{Rich and diverse sampling set}
We call the set of unlabeled documents accessible to the teacher when building models the ``sampling distribution".
We call the set of documents for which models are built the ``deployment distribution''. 
The rich-and-diverse-sampling-set principle is that the sampling distribution captures the richness and diversity of examples in the deployment distribution.
The sampling distribution and the deployment distributions are preferably similar, but they do not have to be perfectly matched. The most important requirement of the sampling distribution is that all important types of documents be represented (rich and diverse). If important documents are missing from the sampling distribution, performance could be impacted in unpredictable ways. As a rule of thumb, unlabeled data should be collected indiscriminately because the cost of storing data is negligible compared to the cost of teaching; we view selectively collecting only the data that is meant to be labeled as both risky and limiting\footnote{For example, the collected set may not contain important examples that would otherwise be found via the machine teaching process.}.
A rich and diverse data set allows the teacher to explore it to express knowledge through selection.
It also allows the teacher to find examples that can be used to train sub-concepts that are more specific than the original concept. For instance, a teacher could decide to build classifiers for bonsai gardening (sub-concept) and botanical gardening (excluded concept) to be used as features to a gardening classifier. The sampling set needs to be rich enough to contain sufficient examples to successfully learn the sub-concepts. The sampling distribution can be updated or re-collected. Examples that have been labeled by teachers, however, are kept forever because labels always retain some semantic value.

\paragraph{Distribution robustness}
The assumption that the training distribution matches the sampling or deployment distribution 
is unrealistic in practice. The role of the teacher is to create a model that is correct for any example, regardless of the deployment distribution. Given our assumption of feature completeness and a rich and diverse sampling set, the result of a successful teaching process should be robust to not knowing the deployment distributions.
Imagine programming a ``Sort" function. We expect ``Sort" to work regardless of the distribution of the data it is sorting. Thanks to realizability, we have the same correctness expectation for teaching. Because the training data is discovered and labeled for the training set in an ad hoc way using filtering, distribution robustness is a critical assumption and we therefore favor machine learning algorithms that are robust to covariate shifts.
Warning: having a mismatch between train and sampling (or deployment) distributions complicates evaluation.

\paragraph{Modular development}
Decomposition is a central principle of both programming and machine teaching. The machine teaching process should support the modular development of concept implementation. This includes the decomposition of concepts into sub-concepts, and the use of models as features for other models. We can achieve this by standardizing model and feature interfaces. Similar to a programming integrated development environment (IDE), within our teaching IDE, concept implementation is done through ``projects" that are grouped into ``solutions". Projects in a solution are trained together because their retraining can affect each other. Dependencies across different solutions are treated as versioned packages, which means that retraining a project in one solution does not affect a project in a different solution (the teacher must update the package reference to incorporate such changes). The modular development principle encourages the sharing of explicit concept implementations. 

\paragraph{Version control} 
All teacher actions (e.g., labels, features, label constraints, schema and dependency graph, and even programming code if necessary) are equivalent to a concept ``program". They are saved in the same ``commit". Like programming code, the teacher's actions relevant to a concept are saved in a version control system. Different type of actions are kept in different files to facilitate merge operations between contributions from different teachers. 

% Algorithm
\begin{algorithm*}[t]
%\SetAlgoNoLine         
\Repeat{forever}{
	\While{training set is realizable}{
    	\If{quality criteria is met} {
      		\textbf{exit}
         }
		\textcolor{blue}{$/ /$ Actively and semantically explore sampling set using concept based filters.}
		
        Find a test error (i.e., an incorrectly predicted (example, label) pair)\;
		Add example to training set.\;
	}
\textcolor{blue}{$/ /$ Fix training set error}

\eIf{training error is caused by labeling error(s)}{
	Correct labeling error(s)\;
	}{
    	\textcolor{blue}{$/ /$ Fix feature blindness. This may entail one or more of the following actions:}
        
		Add or edit basic features\;
		Create a new concept/project for a new feature (decomposition)\;
		Change label constraints or schema (high level knowledge)\;}
	}
\caption{A machine teaching process}
\label{alg:process}
\end{algorithm*}

\vspace{10pt}
The combination of these principles suggests a teaching process that is different from the standard teaching process. The universal teaching language implies that the machine learning expert can be left out of the teaching loop. The featuring completeness principle implies that the engineers can be left out of the teaching loop as well. The teaching tool should provide the teacher with all that is needed to build models effectively. The engineers can update the data pipeline and the programming language, but neither are concept-dependent so the engineer is out of the teaching loop. These two principles imply that a single person with domain and teaching knowledge can own the whole process. The availability of a rich and diverse sampling set means that the traditional data collection for labeling step is not part of the concept teaching process. The distribution robustness principle allows the teacher to explore and label freely throughout the process without worrying about balancing classes or example types. Concept modularity and version control guarantee that a function created in a project is reproducible provided that (1) all of its features are deterministic and (2) training is deterministic. The concept modularity principle enables interpretability and scaling with complexity. The interpretability comes from being able to explain what each sub-concept does by looking at the labels, features, or schema. Even if each sub-concept is a black box inside, their interfaces are transparent. The merge functionality in version control enables easy collaboration between multiple teachers.

% * <gonzo.ramos@gmail.com> 2017-07-16T21:26:37.335Z:
% 
% Is this loop running forever?
% 
% ^.
Based on the above, we propose a skeleton for a teaching process in Algorithm~\ref{alg:process}. Note that this process is not unique. 

Evaluating the quality criteria in a distribution-robust setting is difficult and beyond the scope of this paper. A simple criteria could be to pause when the teacher's cost or time invested reaches a given limit. Finding test error effectively is also difficult and beyond the scope of this paper. The idea is to query over the large sample set by leveraging query-specific teacher-created concepts and sub-concepts. The art is to maximize the semantic expressiveness of querying and the diversity of results. Uncertainty sampling is a trivial and uninteresting case (ambiguous examples are not useful for coming up with new decomposition concepts).

There are a few striking differences between the teaching process above and the standard model building process. The most important aspect is that it can be done by a single actor operating on the true distribution. Knowledge transfer from teacher to learner has multiple modalities (selection, labels, features, constraints, schema). 
The process is a never-ending loop reminiscent of Tom Mitchell's NELL \cite{nell}. Capacity is increased on demand, so there is no need for traditional regularization because the teacher controls the capacity of the learning system by adding features only when necessary. 

%% file: 6-conclusion.tex
\section{Conclusion}

Over the past two decades, the machine learning field has devoted most of its energy to developing and improving learning algorithms. For problems in which data is plentiful and statistical guarantees are sufficient, this approach has paid off handsomely. The field is now evolving toward addressing a  larger set of simpler and more ephemeral problems. While the demand to solve these problems effectively grows, the access to teachers that can build corresponding solutions is limited by their scarcity and cost. To truly meet this demand, we need to advance the discipline of machine teaching. This shift is identical to the shift in the programming field in the 1980s and 1990s. This parallel yields a wealth of benefits. This paper takes inspiration from three lessons from the history of programming. The first one is problem decomposition and modularity. They have allowed programming to scale with complexity. We argue that a similar approach has the same benefits for machine teaching. The second lesson is the standardization of programming languages: write once, run everywhere. This paper is not proposing a standard machine teaching language, but we enumerated the most important machine-learning-agnostic knowledge channels available to the teacher. The final lesson is the process discipline, which includes separation of concerns and the building of standard tools and libraries. This addresses the same limitations to productivity and scaling with the number of contributors that plagued programming (as described in the "Mythical Man Month" \cite{brooks1995mythical}). We have proposed a set of principles that lead to a better teaching process discipline. Some of the tools of programming, such as version control, can be used as is. Some of these principles have been successfully applied in services such as LUIS.ai and by product groups inside Microsoft such as Bing Local. We are in the early stages of building a teaching interactive development environment.

On a more philosophical note, large monolithic systems, as epitomized by deep learning, are a popular trend in artificial intelligence. We see this as a form of machine learning behaviorism. It is the idea that complex concepts can be learned from a large set of (input, output) pairs. With the aid of regularizers and/or deep representations computed using unsupervised or semi-supervised learning, the monolithic learning approach has yielded impressive results. This has been the case in several fields where labeled data is abundant (speech, vision, machine translation). The monolithic approach, however, has limitations when labeled data is hard to come by. Deep representations built from unlabeled data optimize where the data is. Rare misspellings that are domain specific are likely to be ignored or misinterpreted if they appear more frequently in a different domain context. Corner cases with little or no labels for autonomous-driving may be ignored at great perils. Large (amorphic) models are hard to interpret. These limitations can be overcome by injecting semantic knowledge via active teaching (e.g., labels, features, structure). For this reason, we believe that both large monolithic systems and systems more actively supervised by teaching have important roles to play in machine learning. As a bonus, they can easily be combined to complement each other. 

%% file: 7-acknowledgements.tex
\section{Acknowledgements}
We thank Jason Williams for his support and contributions to Machine Teaching. Jason is the creator of the LUIS (Language Understanding Internet Services) project, a service for building language understanding models, based on the principles mentioned in this paper. We thank Riham Mansour and the Microsoft Cairo team for co-building, maintaining, and improving the www.LUIS.ai service. Finally, we thank Matthew Hurst and his team for building high-performing web page entity extractors leveraging our machine teaching tools. These entity extractors are deployed in Bing Local. 
%\end{acknowledgement}